\documentclass{ccjnl}
\usepackage{multirow}
\newtheorem{problem}{Problem}

\title{Digital Twin-Assisted Knowledge Distillation Framework for Heterogeneous Federated Learning}
\author{Xiucheng Wang\inst{1}, Nan Cheng\inst{1,*}, Longfei Ma\inst{1}, Ruijin Sun\inst{1}, Rong Chai\inst{2}, Ning Lu\inst{3}\corinfo{dr.nan.cheng@ieee.org}}
\receiveddate{xxx}
\reviseddate{xxx}
\Editor{xxx}

\address[1]{School of Telecommunications Engineering, Xidian University, Xi'an 710071, China}
\address[2]{Chongqing Key Laboratory of Mobile Communications Technology, Chongqing University of Posts and Telecommunications, Chongqing 400065, China}
\address[3]{Department of Electrical and Computer Engineering, Queen’s University, Kingston K7L 3N6, Ontario, Canada}

\begin{document}
\maketitle

\begin{abstract}
In this paper, to deal with the heterogeneity in federated learning (FL) systems, a knowledge distillation (KD) driven training framework for FL is proposed, where each user can select its neural network model on demand and distill knowledge from a big teacher model using its own private dataset. To overcome the challenge of train the big teacher model in resource limited user devices, the digital twin (DT) is exploit in the way that the teacher model can be trained at DT located in the server with enough computing resources. Then, during model distillation, each user can update the parameters of its model at either the physical entity or the digital agent. The joint problem of model selection and training offloading and resource allocation for users is formulated as a mixed integer programming (MIP) problem. To solve the problem, Q-learning and optimization are jointly used, where Q-learning selects models for users and determines whether to train locally or on the server, and optimization is used to allocate resources for users based on the output of Q-learning. Simulation results show the proposed DT-assisted KD framework and joint optimization method can significantly improve the average accuracy of users while reducing the total delay.
\end{abstract}


\section{Introduction}
Nowadays, FL attracts much attention for researchers since it can train a deep learning (DL) model with high accuracy by indirectly using the whole dataset while preserving data privacy \cite{zhang2021survey}. Since it protects the data privacy of users, FL is widely used for neural network (NN) training in data-sensitive areas such as financial data prediction and medical data analysis. In addition, in FL systems, the user only needs to transmit the parameters of the NN to the server or a specific user to aggregate to improve the performance of the NN, the server does not need to collect data from all users, thus reducing the challenge to the storage capacity of the server when using a large dataset to train a NN with high performance. Moreover, the size of NN parameters is usually much smaller than the size of the training data, so in a poor communication state environment, using FL to only transmit the NN parameters helps to reduce the increase of the total training delay caused by the communication delay \cite{yang2021federated}. The training cost can be further reduced by proper resource management and training methods \cite{shen2021joint,9798090,9718086}.  

However, all the above works assume the NN model in FL is homogeneous, while in the real world, there are various types of devices, and only a few are homogeneous. The different users may have different computing resources and different data even different objectives, so the service needs to be provided in a targeted manner according to the characteristics of the users \cite{9357959,9378794}. The heterogeneity of users leads to many challenges to use the same NN for all users \cite{wang2021device, 9847055, 9488743, 9647925}. Firstly, the heterogeneity of computing resources causes a contradiction between training delay and accuracy in FL systems. Because in FL, the server needs to aggregate the parameters of all users' NNs to improve NN performance, thus if all users train a big model, users with few computational resources from overly dragging down the system training latency. Meanwhile, if all users train a small model, although the training speed can be improved, the accuracy of FL decreases as the model size decreases. Secondly, the heterogeneity of data may cause the data from different users to become not independent and identically distributed (Non-IID). The Non-IID dataset leads to the challenge of convergence of the NN, and the degradation of the NN performance, since in deep learning, the data is generally assumed to follow the IID \cite{li2019convergence}. Another most overlooked but important challenges is the heterogeneity of objectives for users. Obviously, the purpose of user participation in FL systems is to improve the performance of their own NN. However, most previous works focus on the improvement of average accuracy on all users' datasets, which may lead to the degradation of accuracy on its own dataset. As a consequence, FL systems need to improve the average accuracy of the NN on all user datasets while improving the accuracy of the NN on the user's own private dataset.

In practical deployment scenarios, the heterogeneity of computing resources leads to straggler effect in federation learning, which severely reduces training efficiency. This is due to the fact that the computing power of different  devices varies significantly, so that clients with strong computing power have to wait for those with weak computing power before performing model aggregation, resulting in a significant waste of computing resources. Therefore, in \cite{thapa2022splitfed} split learning (SL) is used that users with few computational resources update only a few layers of the NN while other users update more layers of NN. Obviously, the SL increases the communication burden by transmitting the intermediate results of NN between users. 

In addition to the heterogeneity of user computing resources, the heterogeneity of data especially the Non-IID dataset also attracts much attention. Because the heterogeneous data leads to different optimization directions of weights for different models, which is the main reason for the accuracy loss. Therefore, data-based schemes improve model performance by modifying the data distribution. Zhao \emph{et al}. proposed to pre-train the global model by using a small amount of uniformly distributed shared data and distributing the shared data to the client to train the local model \cite{zhao2018federated}. Experiments have shown that even a very small amount of shared data can significantly improve model accuracy. However, this approach of modifying data distribution is often difficult to implement because it is difficult to obtain uniform shared data in practical situations, and constructing shared data increases the risk of user privacy leakage. In addition, users can fine-tune the global model based on client data to obtain a personalized model that performs well on local datasets. Fallah \emph{et al}. proposed to use agnostic meta-learning to obtain the initial global model, and users can obtain high accuracy with a small amount of training based on their own data \cite{fallah2020personalized}. Arivazhagan \emph{et al}. proposed to use personalizing layers in the client model, thus greatly improving the model's adaptability to local datasets \cite{arivazhagan2019federated}. Ma \emph{et al}. achieve personalized model aggregation by discerning the importance of each network layer in different customer models \cite{ma2022layer}. Although these approaches achieve good results on Non-IID data, they all increase the training overhead and do not enable personalizing of the model framework.

To jointly solve the heterogeneity of computing resources, data, and objectives, some works propose to use knowledge distillation (KD) in FL systems. As a promising technology, KD achieves success in many fields, since it can transfer knowledge in a big well-trained NN, named teacher model, to an arbitrary small NN, named student model, by providing the outputs of the big NN as soft labels for student model and using these soft labels as an additional gradient to train the student model. Therefore, the performance of the student model can be improved. Since in the KD method arbitrary NN can be used as a student model, thus different users can select the different models on demand in FL systems. Therefore, by using the KD method, the straggle in the training procedure of FL can be avoided. Because the NNs are not the same between different users, thus there is no need to directly aggregate the NNs of different users, even if the user with more computational resources performs several more training epochs, there is no gradient confusion among different user NNs. Moreover, in KD only the output of the teacher model is needed to transmit thus the transmission burden is much less than SL \cite{singh2019detailed}. Li \emph{et al}. combine migration learning and knowledge refinement for personalized federal learning, and each customer can use private data and a specific model for training. Zhou \emph{et al}. proposed distilled one-shot federated learning (DOSFL), thus significantly reducing communication costs while ensuring model accuracy \cite{li2019fedmd, zhou2020distilled}. Although the KD enables users to select different models based on their demands, however, in KD, a big teacher model is needed, and training such a model usually relay on a huge dataset. Therefore, just using one user's limited dataset cannot train the teacher model well. However. the teacher model is usually too big to train in users' CPU. Thus, it cannot simply use the FL method to train the teacher model. As a consequence, previous work \cite{li2019fedmd}\cite{mothukuri2021survey} mainly relay on the public dataset in the server to train the teacher model in the server. Moreover, if the teacher model is too big, it is challenging for users to run the teacher model to distil the knowledge not only by the limited memory resource to store the huge parameters of the teacher model but also by the large latency to run the teacher model in users CPU.

Fortunately, the evolving digital twin (DT) \cite{8910636} technology can be used to solve these two challenges of no public dataset and users cannot run the teacher model. In DT a digital copy, named twin, all entities and information in physical space is established in a server with adequate computing resources to help improve strategies for physical space \cite{9645156,9860495,9830070,9740161,9579446}. In order to protect the privacy of physical entities, without authorization, only physical entities can access the information of their corresponding twins but not the information of the twins of other physical entities in the digital space, and twins cannot access each other twins' information if they are not authorized by the corresponding physical entities. Therefore, users can safely copy their own datasets and NN parameters to the twin. Moreover, due to the adequate computing resources on the server, the teacher model can be stored and run on the server. In the beginning, the teacher model can be trained as FL by using the data from twins of users. Then users can purchase the computing resource from the server to get the output of the teacher model and thus train their own student model as the KD method. Moreover, to deal with the heterogeneity of users' objectives, each user only uses their own private dataset to train student models. Therefore, not only the average accuracy of student models on all datasets is improved by learning the general knowledge from the teacher model, but also the accuracy of student model on its own dataset is significantly enhanced since only each user's private dataset is used to distillate the knowledge from teacher model. 

Since users may choose to train a small model locally to reduce the cost of purchasing computational resources, as a result, it is important to select appropriate student models for different users to balance the trade-off of the accuracy of NN and the training delay and the training cost, and then based on the selected model determine whether training the student model locally or on the server and purchase appropriate resources to reduce the training delay and training cost. The delay and resource consumption can be effectively reduced by a reasonable scheduling strategy \cite{9097928,8672604,cheng20216g}. Obviously, selecting student models and offloading for all users is a discrete optimization problem, while resource management is a continuous optimization problem. It is challenging for traditional convex optimization to solve such a mixed integer problem. Although Q-learning achieves success in discrete optimization problems and has been proved to always achieve Bellman optimal solutions after training a sufficient number of epochs in discrete optimization problems with limited action space \cite{9380420,9369456}, for problems with continuous action space Q-learning usually performs not well or even fails to converge.

Therefore in this paper, we propose the DT-assisted KD framework to enable different users can select different student models on demand, then a joint optimization method of Q-learning and convex optimization method is used to determine the model selection and offloading and resource management. Remarkably, to couple Q-learning and convex optimization, in the training procedure, the reward of Q-learning is determined by the output of convex optimization. The main contributions of this paper are summarized as follows.
\begin{enumerate}
    \item[1)] A DT-assisted KD framework for heterogeneous FL is proposed. In the proposed framework, users can select different NN models and train models locally or purchase computing resources from the server to train models on demand, to deal with the heterogeneity of computing resources. In the training procedure of user models, only each user's private is used to distill the knowledge from the teacher, thus the user model can not only learn the general knowledge distilled from the teacher model, but also focus on the accuracy on user's private dataset, as a result to deal with the heterogeneity of data and objectives.
    \item[2)]  The joint problem of model selection and training offloading and resource allocation for each user is formulated as a mixed integer programming (MIP) problem. To solve this problem efficiently, a joint optimization method of Q-learning and convex optimization is proposed. The Q-learning is used to select student models and determine whether to train locally or on the server. Based on the decision of Q-learning, the convex is used to allocate resources for users. Remarkably, in the Q-learning training procedure, the objective function value of outputs from convex optimization is used as reward to train Q-learning.
    \item[3)] We evaluate the performance of the proposed method through extensive simulations. The results show that the proposed FL framework achieves significantly higher performance than the typical FL method in heterogeneous systems. The results also show the proposed joint optimization method can reduce the delay in the FL training procedure and the cost of transmission and computing resources while ensuring the student model performance for each user. 
\end{enumerate}

The remainder of the paper is organized as follows. In Section \ref{sec-2} the proposed KD-driven FL training framework assisted by DT is described, and the problem of selecting models for all users and resource management is formulated. Section \ref{sec-3} describes the convex optimization assisted Q-learning method to solve the problem in Section \ref{sec-2}, and the details of the KD training method are described. Section \ref{sec-4} evaluates the performance of the proposed method and Section \ref{sec-5} concludes the paper.
\begin{figure*}[h]
  \centering
  \includegraphics[width=1.4\columnwidth]{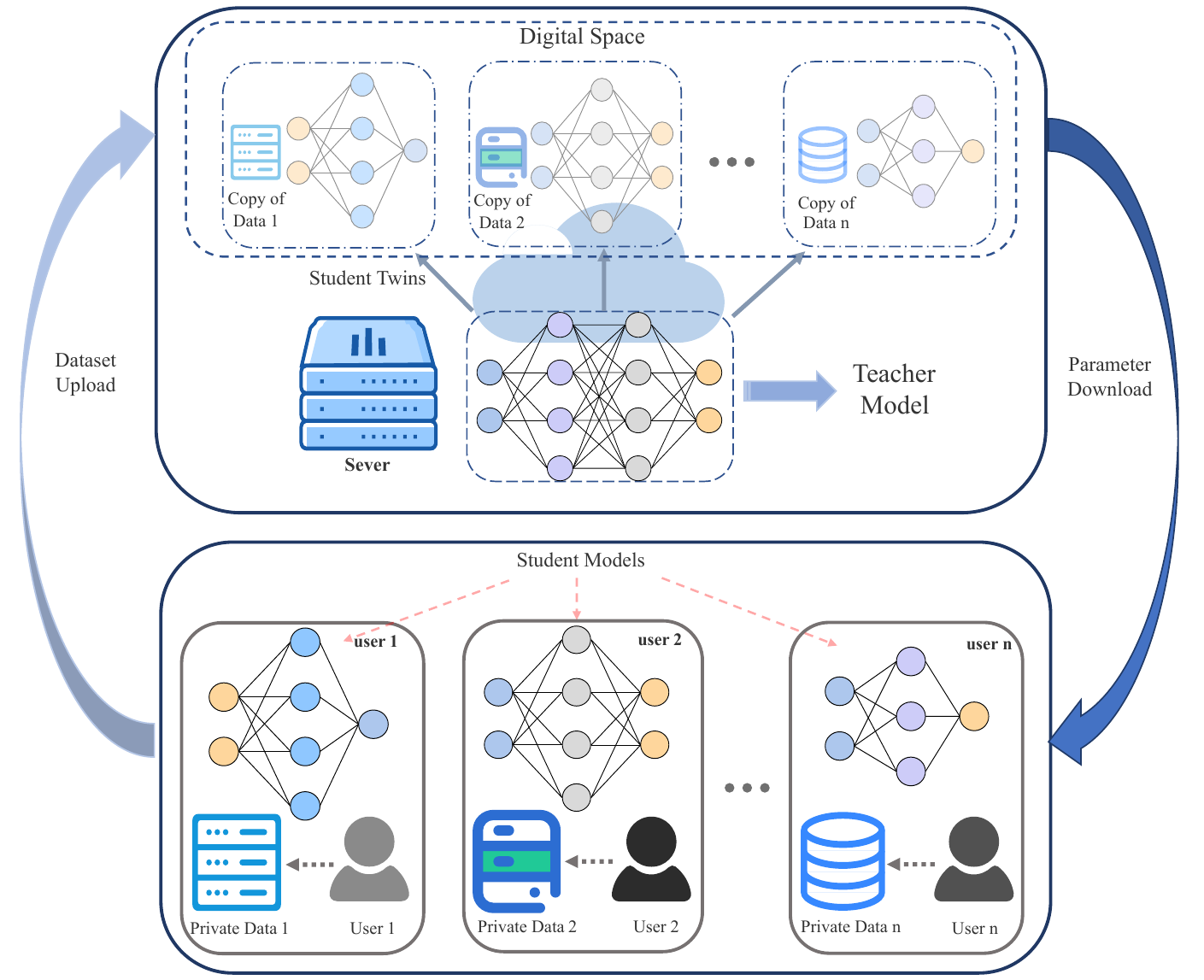}

  \centering \caption{\textbf{DT-assisted KD framework in FL.} A powerful teacher model is trained in the digital space using DT, and later a copy of the teacher is used for distillation for each student model. During training, the teacher model is only located in the digital space, while the student model is located in the physical space on the user side and there is a copy of student model in digital space. When computing the student model in the physical space, the feature labels of the teacher model need to be accepted periodically and the twins in the digital space need to be synchronized. \textbf{User private data can only be accessed locally or in the twin and cannot be shared.}}
  \label{fig_overview}

\end{figure*}

\section{System Model and Problem Formulation}\label{sec-2}
\subsection{System Framework}


As shown in Figure~\ref{fig_overview}, the proposed DT-assisted KD framework for heterogeneous FL system is consisted by the two space: digital space and physical space. The physical space including $N$ users, and each user $i$ equips a central processing unit (CPU) with frequency $f_{loc,i}$ and a private dataset $\mathcal{D}_i$, the aim of each user $i$ is to efficiently train a student NN model of size $\theta_{s,i}$ with high accuracy and low costs. In digital space, there is a big teacher model of size $\theta_t$ and the $K$ twins for users, where each twin of user has a copy of user dataset and the parameters of user student model, and the digital space is run on a server with CPU frequency as $f_{ser}$. Before training, the teacher model is firstly trained by the copy of dataset in digital space through a FL method to protect the privacy of dataset copy. Then in each epoch, the user model is trained on its private dataset to distill the knowledge from teacher model, thus improving the performance on student model. Assuming the $f_{ser}$ is much larger than any $f_{loc,i}$ and user can purchase the computing resources from the server, user can train the student model locally or on the server. If the student model is train locally, the server should transmit the output of teacher to user that the knowledge in teacher model can be distilled. Otherwise, the digital copy of student model is trained on the server, and the transmission of output from teacher can be neglect, since they are both in the same server. To ensure the model parameters are consistent in digital and physical space, therefore, after each training epoch the digital/physical space should transmit the parameters of student to physical/digital space, depending on training the student model locally or on the server.


Because  $f_{loc,i}$ can be various, simply using the same NN will cause the contradictory between accuracy and training delay. To deal with such contradictory, a KD is used to share the knowledge between different NNs. Since in KD, the knowledge from teacher can be distilled to student model with arbitrary depth and width, as a result, different users can choose the different size of NNs on demand. In order to emphasize the impact of different NN sizes on FL system performance and computational latency, in this paper, instead of designing NN network depth and width specifically as in neural architecture search (NAS) \cite{elsken2019neural}, several representative neural network architectures are selected to form a set $\mathcal{M}$, each element in $\mathcal{M}$ represents a NN with a specific depth and width, and users can select NNs on demand in $\mathcal{M}$.

To deal with heterogeneity of data and objective, the teacher model is first trained by the data copy in digital space of all users in a FL method, thus the teacher model can extract the general knowledge from all dataset. Then in the training procedure of each user model, only the user private dataset is used to distill the knowledge from teacher model. Therefore, the user model can not only learn the general knowledge of all users dataset from teacher model, but also focus on learn the personalized knowledge from the user private dataset, since only private data is used in training. 

Generally, in KD, the big teacher model is well-trained in advance. However, due to the heterogeneity of user computing resource, in this paper, we assume the teacher model is too big to train by the user. Inspired by the feature of digital twin (DT), a DT-assisted KD framework is proposed to enable the training of teacher model and selecting different model for different users. According to \cite{liu2021review}, the digital twin system can establish a digital copy, named a twin, of the total entities and data in physical space. Therefore, users can create a data backup in the digital space. Meanwhile, to protect the privacy of user data, twins in the digital space are not allowed to exchange the user's private data with other twins or share data with other users in the physical space. However, in order to share the data indirectly, using the data to train NN is allowed in the twin, since in the training procedure the data is always stored in the twin and when the training is finished other users and twins can only get a NN with trained parameters or gradient. Therefore, the teacher model can be trained on the total dataset to get high performance. Then the KD can be used and the teacher model can provide an additional gradient to the small NN, called the student model, in the user, thus improving the performance of the student model.

In this paper, to emphasize the importance of model selection and resource management for different users, we consider that the DT system is pre-built and that the twin in DT already has a replica of the user's data, and the teacher model has been well trained. According to \cite{gou2021knowledge}, in the KD system, the student model needs to update its parameters by using the calculation outputs or intermediate results of the teacher model. However, since the teacher model is too big to store in the user, the teacher model is always stored in the server, thus the user needs to purchase the computing resource from the server to get the outputs of the teacher model. Assuming the server equips a CPU with adequate computing resource as $f_{ser}$, and the $f_{ser}$ is much larger than any CPU frequency of users. In order for the server to provide services to all users at the same time, the computing resources of the server are therefore divisible and the server can communicate with all users at the same time. Assuming that the server communicates with users by a wireless network, thus inspired by the success of orthogonal frequency division multiplexing (OFDM) in 4G and 5G networks, the OFDM is used in this system to avoid interference among users with maximum bandwidth $b_{max}$. Therefore, the transmission rate between user $i$ and server is
\begin{align}
    r_i = b_i \log_2(1+\frac{p_ih_i}{n_0}),
\end{align}
where $b_i$ is the bandwidth used to transmit information between the server and user $i$, $h_i$ is the channel gain between user $i$ and server, $p_i$ is the transmission power for user $i$, and $n_0$ is the noise power.  

In order to fully utilize both user and server computing power, users can choose to train their own student models by using local computing power or by the server computing power on demand. Regardless of which method a user chooses to train the student model, the training overhead is composed of both latency and resource price. If the user trains the student model by using its own computing power, in the training procedure, the user needs to wait for the server to transmit the results calculated by the teacher model to the user first. Then the user uses the results of the teacher model and the labels of the dataset to update the parameters of the student model locally by the back-propagation algorithm. Since the features of the twin in the digital twin system need to be guaranteed to be consistent with the features of the entities in the physical space, the user also needs to transmit the updated parameters of the student model to the server to ensure that the features of the entities in the digital space and the physical space are consistent. Therefore, the delay in training the student model locally consists of computational latency of the teacher model $T_{tea}$, transmission delay for users to obtain the computing results of the teacher model $T_{label}$, computational delay for updating student model parameters $T_{stu}$, and the transmission delay of synchronized model parameters in physical and digital space $T_{model}$. According to \cite{lecun2015deep}, for an NN with a determined depth and width, the computation time is inversely proportional to the CPU frequency. Therefore, if user $i$ purchases computing resources of size $f_i$ from the server, the delay of the computing of teacher model parameters updating of student model in user $i$ and is
\begin{align}
    &T_{tea,i} = \frac{\mu_t}{f_{i}},\\
    &T_{stu,i} = \frac{\mu_{s,i}}{f_{loc,i}},
\end{align}
where $\mu_t$ and $\mu_{s,i}$ is the time of teacher computing and the time of updating the parameters of student model by using unit of computing resources. Assuming the size of output of teacher model and parameters of student model in user $i$ is $\theta_l$ and $\theta_{s,i}$, the transmission delay 
\begin{align}
   & T_{label,i} = \frac{\theta_l}{r_{i}},\\
   & T_{model,i} = \frac{\theta_{s,i}}{r_{i}}.
\end{align}
If the user chooses to train the student model using the server arithmetic, then since both the teacher model and the student model are on the server, the delay in transmitting the teacher model output is not included in the total latency, after the parameters of the student model are updated at the server, they need to be downloaded to the local area, so the transfer delay includes only $T_{model,i}$. Moreover, the parameter update time of the student model becomes $T_{stu,i} = \frac{\mu_{s,i}}{f_{i}}$. 

\subsection{Problem Formulation}
To formulate the problem we define two discrete optimization variables $\mathbf{x}$ and $\mathbf{m}$. The $\mathbf{x}$ denotes whether the student model is trained locally or on the server, and the $\mathbf{m}$ denotes which student model selected by user. Above all, the problem in this paper can be formulated as following.
\begin{problem}
\begin{align}
&\min_{\mathbf{m,x,f,b}}\quad\sum_{i=1}^{N} \alpha(T_{tea,i}+T_{stu,i}+T_{model,i}+x_iT_{label,i}) \notag\\
    & \qquad\qquad\qquad+ \beta \left(T_{tea,i}f_i + \left(1-x_i\right)T_{stu,i}f_i\right) + \delta b_i\notag\\
    & \qquad\qquad\qquad- \eta_o Acc_{own,i} - \eta_a Acc_{avg,i}
    ,\label{obj}\\
&s.t.\;\; x_i \in \{0,1\},\label{c1}\\
&\qquad m_i \in \{1,2,\cdots,|\mathcal{M}|\},\label{c2}\\
&\qquad \sum_{i=1}^{N} b_i \le b_{max},\label{c3}\\
&\qquad \sum_{i=1}^{N} f_i\le f_{ser},\label{c4}\\
&\qquad T_{tea,i} = \frac{\mu_{t}}{f_i},\label{c5}\\
&\qquad T_{stu,i} = \left\{\begin{array}{l}
             \mu_{m_i}/f_i \qquad\; x_i = 0, \\
             \mu_{m_i}/f_{loc,i}  \quad x_i =1.
             \end{array}\right.\label{c6}\\
&\qquad r_i = b_i \log_2(1+\frac{p_ih_i}{n_0}),\label{c7}\\
&\qquad T_{label,i} = \frac{\theta_l}{r_{i}},\label{c8}\\
&\qquad T_{model,i} = \frac{\theta_{s,m_i}}{r_{i}},\label{c9}\\
&\qquad Acc_{own,i} = \frac{1}{|D_i|} \mathbb{I}(y=\text{NN}_{m_i}(in)),\label{c10}\\
&\qquad Acc_{avg,i} = \sum_{i=1}^{N}\frac{1}{|D_i|} \mathbb{I}(y=\text{NN}_{m_i}(in)),\label{c11}
\end{align}
\end{problem}
where $\alpha$ $\beta$ $\delta$, and $\eta_o$ $\eta_a$ are constant factors, $\mu_{m_i}$ is the time to updating the parameters of student model in user $i$ using unit computing resource. The first part in Equation (\eqref{obj}) is the total delay to train the student model and when the student model is trained locally, the $x_i$ is 0 thus the transmission delay of the teacher model output $T_{label,i}$ is 0. The second part is the cost of purchasing the computing resource from the server, and when the student model is trained locally the $\left(1-x_i\right)T_{stu,i}f_i$ is 0, which means the user only pays for the computing resource to get the output of teacher and the cost of bandwidth, the third part of Equation (\eqref{obj}) is the accuracy of the student model tested on its own dataset and the overall datasets of all users, and since we want to maximize the accuracy of model the negative values of accuracy is minimized. The Constraint(\eqref{c1}) means the student model can be trained either locally or on the server. Constraint (\eqref{c2}) ensures the user can choose the student model only from the model set $\mathcal{M}$. Constraints (\eqref{c3}) and (\eqref{c4}) show the sum of computing resources and bandwidth resources purchased by all users cannot exceed the maximum compute resources and maximum bandwidth resources. Constraint (\eqref{c5}) is the computing delay of the teacher model, while the training delay of student model in constraining (\eqref{c6}) depend on the whether it is trained by user or server. The constraint (\eqref{c7})-(\eqref{c9}) is the transmission delay of teacher output and parameters of the student model. Constraint (\eqref{c10})-(\eqref{c11}) is the method to evaluate the accuracy of the student model, where $\mathbb{I}(*)$ is the indicator function, NN$_{m_i}$ is the student model of user $i$, $in$ is the input for a sample of dataset and $y$ is the label of the sample.
\begin{figure}[h]
  \centering
  \includegraphics[width=1\columnwidth]{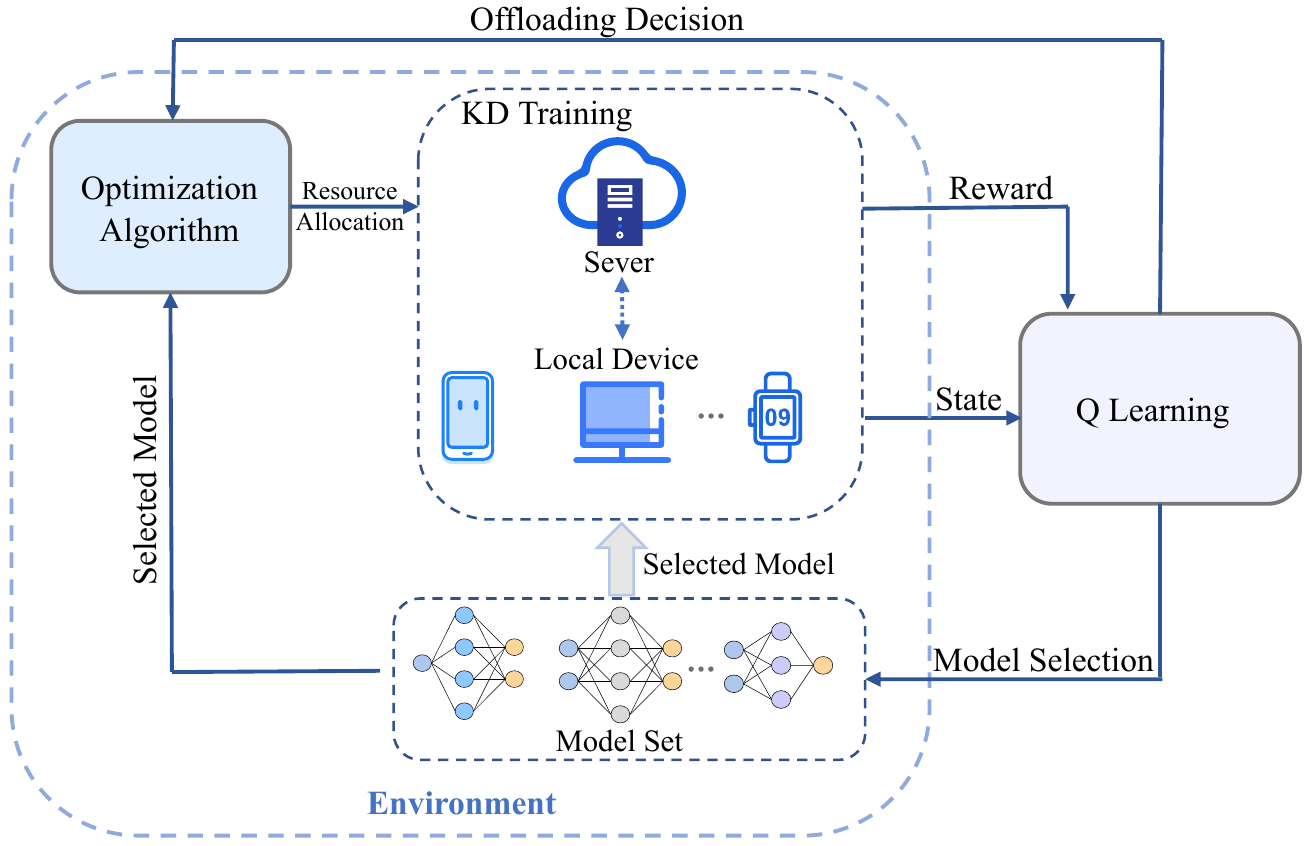}

  \centering \caption{\textbf{Overview of the proposed method.} It consists of three parts: KD-driven model training, RL-based model selection and offloading, and optimization-based resource allocation. Q learning is responsible for selecting customer models and making offloading decisions, while later convex optimization is used to allocate resources for users. After the model training is completed using KD, the reward is obtained.}
  \label{method-q}

\end{figure}

\section{Proposed Method}\label{sec-3}
\subsection{Convex Optimization Assisted Q-learning Method}

Obviously, Problem 1 is a mixed integer programming (MIP) problem, thus it is challenging to use the traditional convex optimization method to optimize the problem with discrete variables efficiently. Since we can simultaneously select student models for all users, in this way, selecting student models for the current users does not have an effect on the selection of student models for other users. Therefore, similar to \cite{meng2019power}, the student model selection can be formulated as a Markov decision procession (MDP) without state transition probability.
Although Q-learning achieves success in discrete optimization problems both in efficiency and performance, it is challenging for it to solve the problem with continual optimization variables. Unfortunately, in Problem 1, the $\mathbf{x,m}$ are discrete while the $\mathbf{f,b}$ are continual, thus, this problem is challenging to be solved by using convex optimization or Q-learning alone. Although there are many deep neural networks (DNN)-based reinforcement learning methods, such as deep deterministic policy gradient (DDPG) \cite{ddpg} and proximal policy optimization (PPO) \cite{ppo}, which can jointly optimize the discrete and continual variables, the training of the DNN can be time-consuming. It is unnecessary to train a DNN-based RL first, and then use this DNN model to optimize the training of FL. Moreover, the inferencing time of DNN-based methods to get the optimized variables is also long, which cannot be simply ignored in delay-sensitive tasks \cite{9882279}, thus increasing the total processing time. As for the Q-learning, we can further store the Q-Table as a Hash-Table. In this way, we can get the solution from Q-learning with computing complexity as $\mathcal{O}(1)$ \cite{maurer1975hash}. As a result, the latency caused by the optimization can be ignored. Therefore, in this paper, we use the Q-learning-based method to optimize the problem. Fortunately, further analysis of Problem 1 reveals that the optimization problem for $\mathbf{f}$ and $\mathbf{b}$ given $\mathbf{x}$ and $\mathbf{m}$ is a convex problem, and both $\mathbf{x}$ and $\mathbf{m}$ are discrete variables that can be optimized using Q-learning. Thus Problem 1 can be decomposed into two subproblems: 1) choosing a student model for each user and determining whether it is trained with user arithmetic or server arithmetic, which can be solved by Q-learning; and 2) assigning appropriate computational and bandwidth resources to each user, which can be solved by convex optimization. Since both Q-learning and convex optimization can get the solution quickly, the influence of the inferencing time of the algorithm can be ignored. As a consequence, Q-learning is used to solve the problem as follows.
\begin{problem}
\begin{align}
&\min_{\mathbf{m,x}}\quad\sum_{i=1}^{N} \alpha(T_{tea,i}+T_{stu,i}+T_{model,i}+x_iT_{label,i}) \notag\\
    & \qquad\qquad\qquad+ \beta \left(T_{tea,i}f_i + \left(1-x_i\right)T_{stu,i}f_i\right) + \delta b_i\notag\\
    & \qquad\qquad\qquad- \eta_o Acc_{own,i} - \eta_a Acc_{avg,i}
    ,\label{reward}\\
&s.t.\;\; x_i \in \{0,1\},\\
&\qquad m_i \in \{1,2,\cdots,|\mathcal{M}|\},\\
&\qquad\text{constraint } (\eqref{c5})-(\eqref{c11}),
\end{align}
\end{problem}
The relevant function for Q-learning is defined as

\textbf{Reward:} We expect the optimization algorithm to obtain the optimal objective function value after the action selection is completed by Q learning, thus the Equation (\eqref{reward}) is used as the reward function $r$.

\textbf{State:} The difference in arithmetic power of each user device leads to different speeds of training. In addition, the channel state between the client and the server affects the data transmission rate. In order to obtain the largest possible system gain, a balance between latency and accuracy needs to be found according to user characteristics. Therefore, the state $s=\left\{\mathbf{f}_{\text {loc }}, \mathbf{h}\right\}$, where $\mathbf{f}_{\text {loc}}$ and $\mathbf{h}$ are the vectors of CPU frequency and channel state, respectively.

\textbf{Action:} Since Q learning performs model selection and training scheduling, the action $a=\left\{\mathbf{x}, \mathbf{m}\right\}$.

Based on this, the Q table update rule can be expressed as
\begin{align}
\begin{split}
Q\left(s, a\right) \leftarrow  &Q\left(s, a\right)\\ 
&+\alpha\left[r+\gamma \max _{a^{\prime}} Q\left(s^{\prime}, a^{\prime}\right) - Q\left(s, a\right)\right], \label{update} \\
\end{split}
\end{align}
where $s^{\prime}$, $a^{\prime}$ denote the state and action information of the next moment, $\alpha$, $\gamma$ are the learning rate and fading factor, respectively.

Then based on the optimized variables $\mathbf{x}$ and $\mathbf{m}$ by Q-learning, the subproblem of optimizing the $\mathbf{f}$ and $\mathbf{b}$ is a convex optimization problem as follows.

\begin{problem}
\begin{align}
&\min_{\mathbf{m,x}}\quad\sum_{i=1}^{N} \alpha(T_{tea,i}+T_{stu,i}+T_{model,i}+x_iT_{label,i}) \notag\\
    & \qquad\qquad\qquad+ \beta \left(T_{tea,i}f_i + \left(1-x_i\right)T_{stu,i}f_i\right) + \delta b_i,\label{obj3}\\
&s.t.\;\; \qquad\text{constraints } (\eqref{c3})-(\eqref{c9}).
\end{align}
\end{problem}

Different from Problem 1 and Problem 2 the objective function (\eqref{obj3}) of the performance of NN is not included. This is because the performance of each user's NN is only related to the structure of the selected student model, the optimization of computational and bandwidth resources only affects the speed of training and does not have an impact on the performance of NN. Obviously, Problem 3 is convex thus we can solve it by convex optimization.

Therefore, as is shown in Figure \ref{method-q} the Q-learning first selects student models for all users and determines whether the student model is trained locally or on the server. Then, based on the selected student models and offloading decision, convex optimization is used to allocate the resources for all users. Then the value of the objective function (\eqref{obj}) can be calculated, which is used as the reward to train the Q-learning.

\subsection{Teacher and Student Model Training Method Assisted by DT}
Knowledge distillation (KD) is a neural network model compression technique based on teacher-student network structure. It has been widely used recently due to its excellent performance and high compatibility with other deep learning techniques. Usually, there are differences in the performance of models of different scales. Models with more parameters and high computational complexity are able to obtain higher accuracy than smaller models. However, due to the limitations of device computing and storage capacity, as well as specific task requirements, it is not possible to deploy large models in many cases. Therefore, KD is used to compress the model while minimizing the loss of accuracy, and can also be considered as a way to use a large model to improve the performance of a small model. 

To apply KD, a robust teacher model is necessary. Training a teacher model requires powerful computing power and memory that cannot be satisfied by the client. Therefore, we use digital twin technology to copy user data to the digital domain and train the teacher model using the server. We use FedSGD \cite{mcmahan2016federated} to train the teacher model, and aggregate the gradients to update the global model after each user finishes computing on a batch of data. Due to the high heterogeneity of users' private data, we set the batch size to the size of the private dataset to enhance the stability and accuracy of model convergence. As a result, the full user data is used for each parameter update, thus avoiding the performance degradation caused by aggregating the model parameters in different optimization directions. It should be noted that although this approach can obtain a highly accurate teacher model with guaranteed data security, it consumes a large amount of computing power and storage space.

Traditional KD methods use the output of the teacher model as a soft target to help optimize the parameters of the student model. Specifically, the loss function of the student network can be expressed as 
\begin{equation}
\begin{aligned}
     L_{\text{KD}}=L_{\text{hard}} + T^{2}L_{\text{soft}} ,\label{Lkd}
\end{aligned}
\end{equation}
where $T$ is the temperature coefficient. $L_{\text {hard}}$ denotes the traditional cross-entropy loss, and $L_{\text {soft}}$ is the soft target obtained from the prediction pairs of the teacher model and student model
\begin{equation}
\begin{aligned}
&L_{\text {hard}}=L_{\text{CE}}\left(\boldsymbol{p}_{s}, \boldsymbol{y}\right)\\
&L_{\text {soft}}=L_{\text{KL}}\left(\boldsymbol{p}_{s}, \boldsymbol{p}_{t}\right) ,
\end{aligned}
\end{equation}
where $L_{\text{CE}}(\cdot)$ is the cross-entropy function, $L_{\text{KL}}(\cdot)$ denotes the Kullback-Leibler divergence, $\boldsymbol{y}$ is the label vector, and $\boldsymbol{p}_{s}$, $\boldsymbol{p}_{t}$ are the prediction vectors of the student model and the teacher model, respectively. Compared with the traditional single-model training method using only the hard loss function, the soft target provided by the teacher model provides more knowledge for the student model, thus improving the accuracy of the student model.

Despite the great success of KD in many areas, there is still a significant gap between the performance of the teacher model and the student model, and the performance improvement of the student model relies heavily on the appropriate model framework and the rationality of the hyperparameter values. To further improve the performance of student models, several advanced distillation schemes have been proposed. These schemes perform deeper knowledge extraction and employ more refined knowledge representation methods\cite{wang2021knowledge,chen2021cross,komodakis2017paying}. However, they have stringent requirements on the model framework and their training process tends to be more complex. Distillation of multiple significantly different student models using the same teacher model is difficult to achieve satisfactory performance while training multiple teacher models for different student models incurs significant overhead. Moreover, for complex tasks with high training difficulty, the gains obtained by relying only on soft targets for distillation are insufficient. Therefore, we use a more powerful and simple distillation method (simKD) that reuses the linear output layer of the teacher model for deeper knowledge extraction\cite{chen2022knowledge}.

Usually, the classification model relies on an encouder consisting of several hierarchical structures connected by elaborate design for feature extraction, thus transforming low-level features of images into high-level features and finally outputting classification results through linear layers. This feature extraction capability is required by the model to process a variety of images so that a model trained on a specific dataset can be migrated to other datasets with only fine-tuning of the parameters to achieve good performance. The classification accuracy of a model depends heavily on its feature extraction capability, so most of the existing work has focused on designing more efficient feature extraction structures. For different models, most of their computational effort comes from the feature extraction part, while the linear layer used as the classifier accounts for only a small percentage. Therefore, in order to make the student model approximate the performance of the teacher model as much as possible, we match the student model with the teacher model in terms of features, and the objective function of the student model is defined as
\begin{equation}
\begin{aligned}
     L_{\operatorname{simKD}}=\left\|\boldsymbol{f}_{t}-P\left(\boldsymbol{f}_{s}\right)\right\|^{2},
\end{aligned}
\end{equation}
where $\boldsymbol{f}_{t}$, $\boldsymbol{f}_{s}$ denotes the feature vectors of the teacher model and the student model, respectively, and $P(\cdot)$ is a projector to match the dimensionality of the feature vectors. To further improve the accuracy of the student model and reduce the training overhead, the student model is made to directly reuse the linear layer of the teacher model to ensure that it can perform correct classification based on the learned feature knowledge. Since the linear layers of our chosen models have the same or similar structure, the computational complexity of the student model remains almost unchanged.

\begin{figure}[h]
  \centering
  \includegraphics[width=0.9\columnwidth]{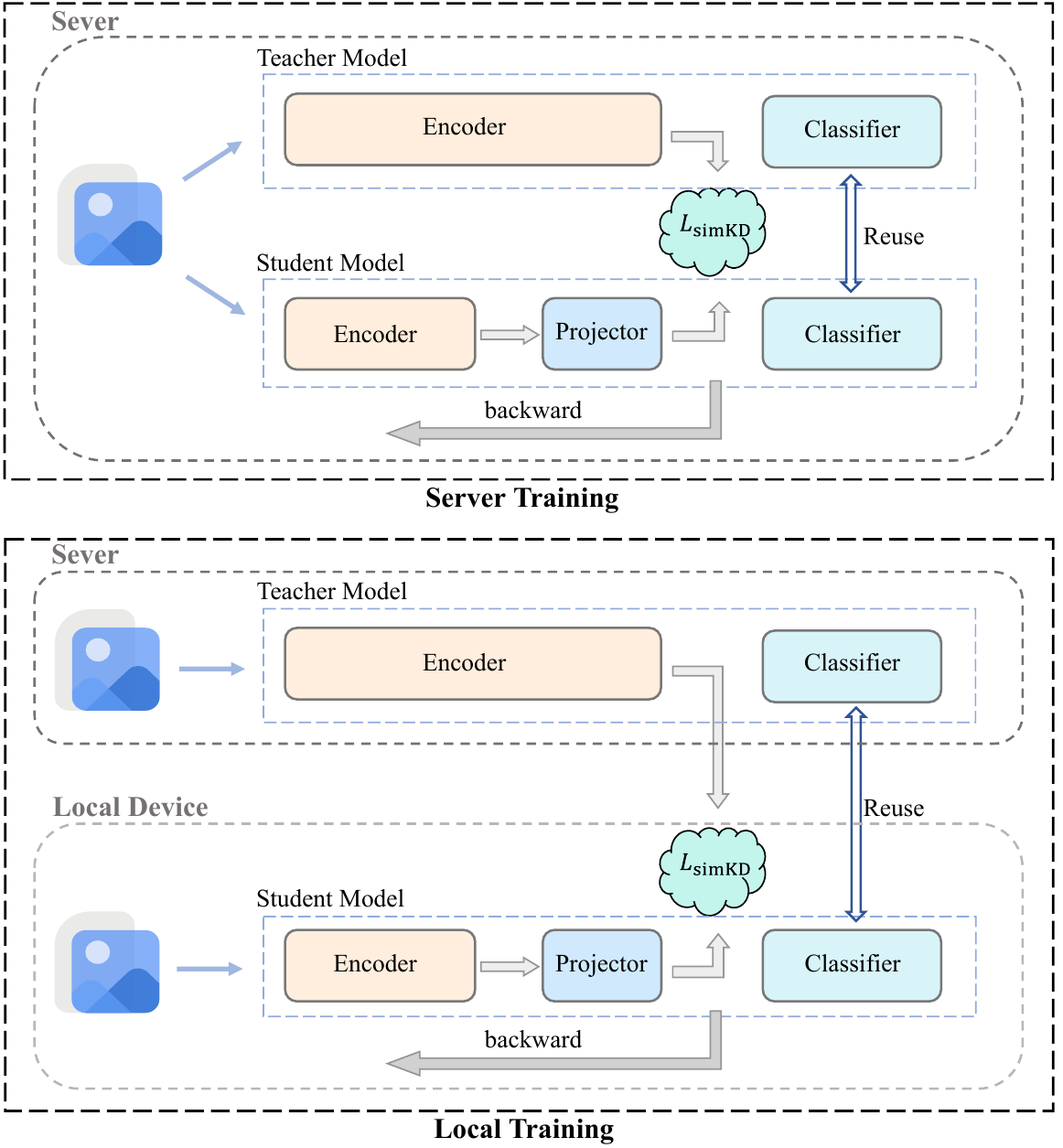}

  \centering \caption{\textbf{Training management.} There are two scenarios for model training: training on the server and training locally, depending on the location where the \textbf{student} model is computed.}
  \label{method-kd}

\end{figure}

Using the client model as the student model, we show in Figure \ref{method-kd} the process of training the client model using simKD at the server and at the local side, respectively. When training on the server, the teacher model and the client model are fed with the same data, and the features of the teacher model are used as label pairs for back-propagation of the client model. When training locally, the forward propagation of the teacher model is done on the server, and then the feature labels are sent to the user, and the forward and backward propagation of the client model is done locally.

\section{Simulation Results}\label{sec-4}
In this section, we evaluate the performance of the proposed DT-assisted KD framework and the joint resource management method of the Q-learning convex optimization method in simulation. The transmission power $p$ of all users is 0.1w, and the maximum bandwidth $b_{max}$ is 10 Mhz. All users are randomly distributed within a distance $d$ of 10 to 100 meters from the server, and the channel gain $h=\frac{g_0}{d^{\gamma}}$ with $g_0=-40$ dB and $\gamma=2.8$. The CPU frequency is 10 GHz and that of users is randomly in $[0.5, 2]$ GHz. The number of users $K$ is 4 in the simulation. To demonstrate the ability of the proposed DT-assisted KD training method (KD) to adapt to heterogeneous data and devices, we choose four models of different sizes to match the client devices. Specifically, VGG-8, ResNet-8x4, ResNet-14x4, and ResNet-26x4 are used as student models and ResNet-32x4 as the teacher model, and the performance is tested on datasets with different distributions. The time of training student model for one epoch with unit giga CPU frequency is 6.83s, 8.75s, 12.27s and 18.96s respectively.

\textbf{Datasets and baselines.} We used two ways to slice the CIFAR-100 dataset: Images from all categories are divided into four equal parts to form four IID sub-datasets. Divide the dataset according to image categories, each having 25 categories of images, to obtain four mutually orthogonal Non-IID sub-datasets. We compare the proposed algorithm with traditional federal learning\cite{mcmahan2017communication} (FL), as well as with the way to train student models individually at each client (STU). It is noted that among the three algorithms, the KD-driven method and the student-only model training approach are able to deploy suitable models for different devices, while the conventional FL is able to use only one model due to the need for parameter aggregation.

\textbf{Training details.}
We use the SGD optimizer for training and set the training epoch to 100. For KD, we set the Nesterov momentum and initial learning rate to 0.9 and 0.05 and divide the learning rate by 5 at the 45th, 60th, 75th, and 90th of the total 100 epochs. The mini-batch is set to 64 and the weight decay is set to $7.5 \times 10^{-4}$.
For FL, we set the parameters based on \cite{mcmahan2017communication}. STU uses the same parameters as KD.

\subsection{Performance of DT-Assisted KD Training Method}

\begin{figure}[h]
  \centering
  \includegraphics[width=0.9\columnwidth]{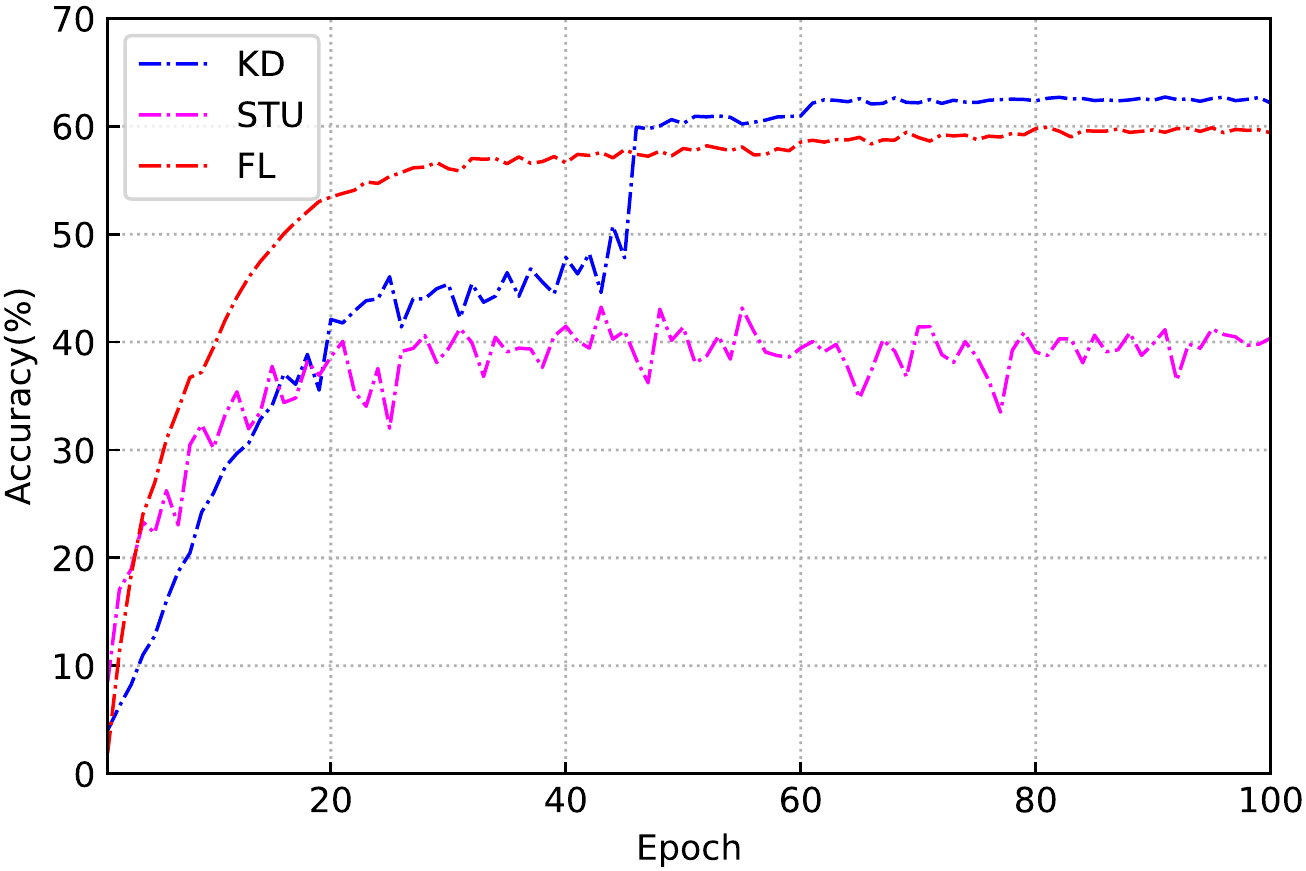}

  \caption {Test accuracy with IID training set.}
  \label{fig2}

\end{figure}

\begin{figure}[h]
  \centering
  \includegraphics[width=0.9\columnwidth]{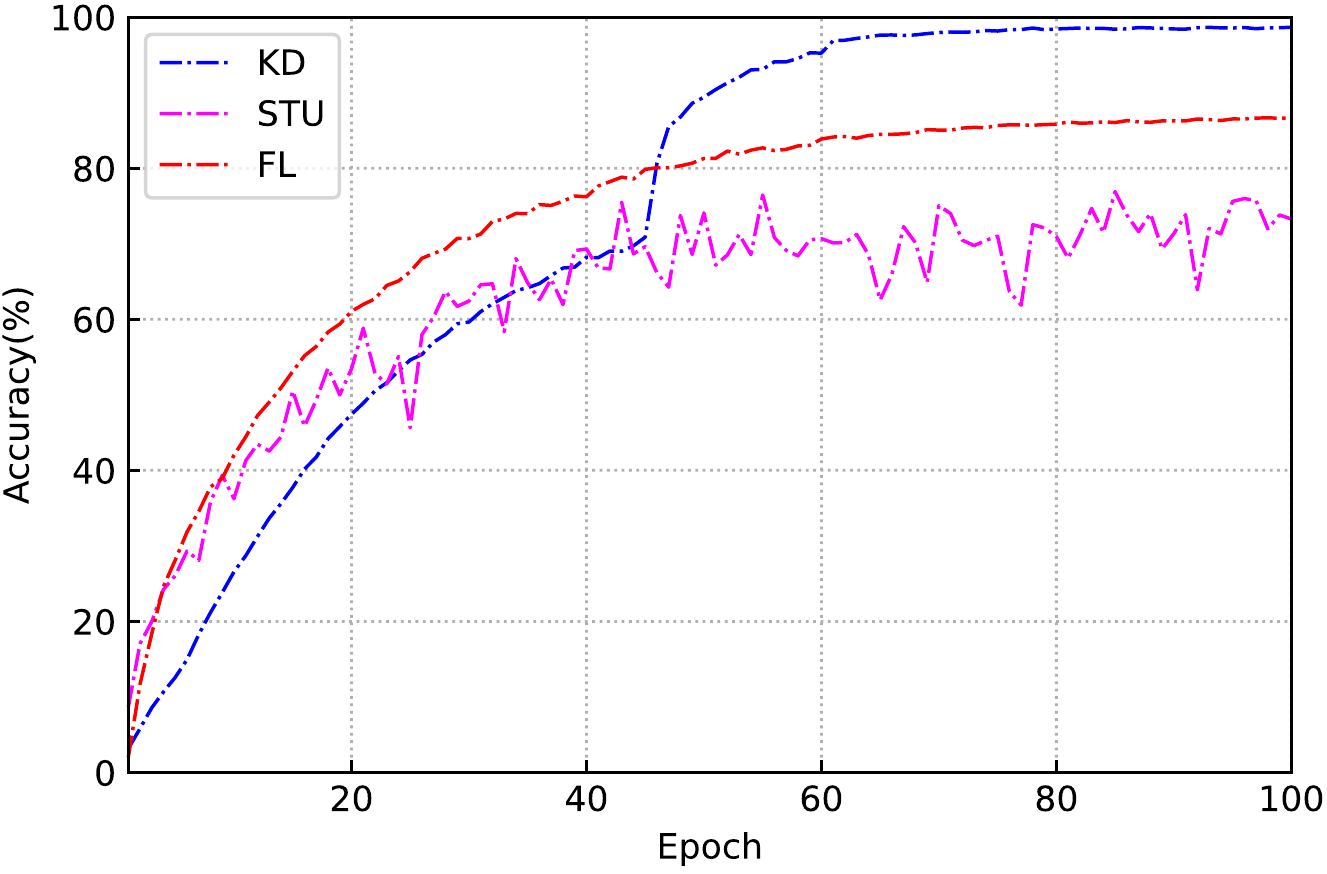}

  \caption {Train accuracy with IID training set.}
  \label{fig3}

\end{figure}

To show the performance difference between the different methods, we compare the accuracy obtained by training the vgg-8 on different datasets. Figure \ref{fig2} and Figure \ref{fig3} show the accuracy curves of each algorithm trained on the IID dataset. It can be observed that both KD and conventional FL achieve good performance, while the accuracy of the student model trained alone is poor. This is due to the fact that the first two methods are able to rely on a strong teacher model or other student models to update their own parameters and broaden the knowledge sources of the model, thus effectively improving the model performance. In contrast, training student models based on their own data only leads to a severe limitation in model accuracy due to the lack of training data volume. Specifically, KD and FL were able to obtain $62.71\%$ and $59.90\%$ accuracy, respectively, significantly better than STU's accuracy limit of $43.23\%$, and their convergence stability is also seen by the images to be better. Although the conventional FL can achieve good results on the IID dataset, the KD-driven training method still achieves an accuracy improvement of $2.81\%$.

\begin{figure}[h]
  \centering
  \includegraphics[width=0.9\columnwidth]{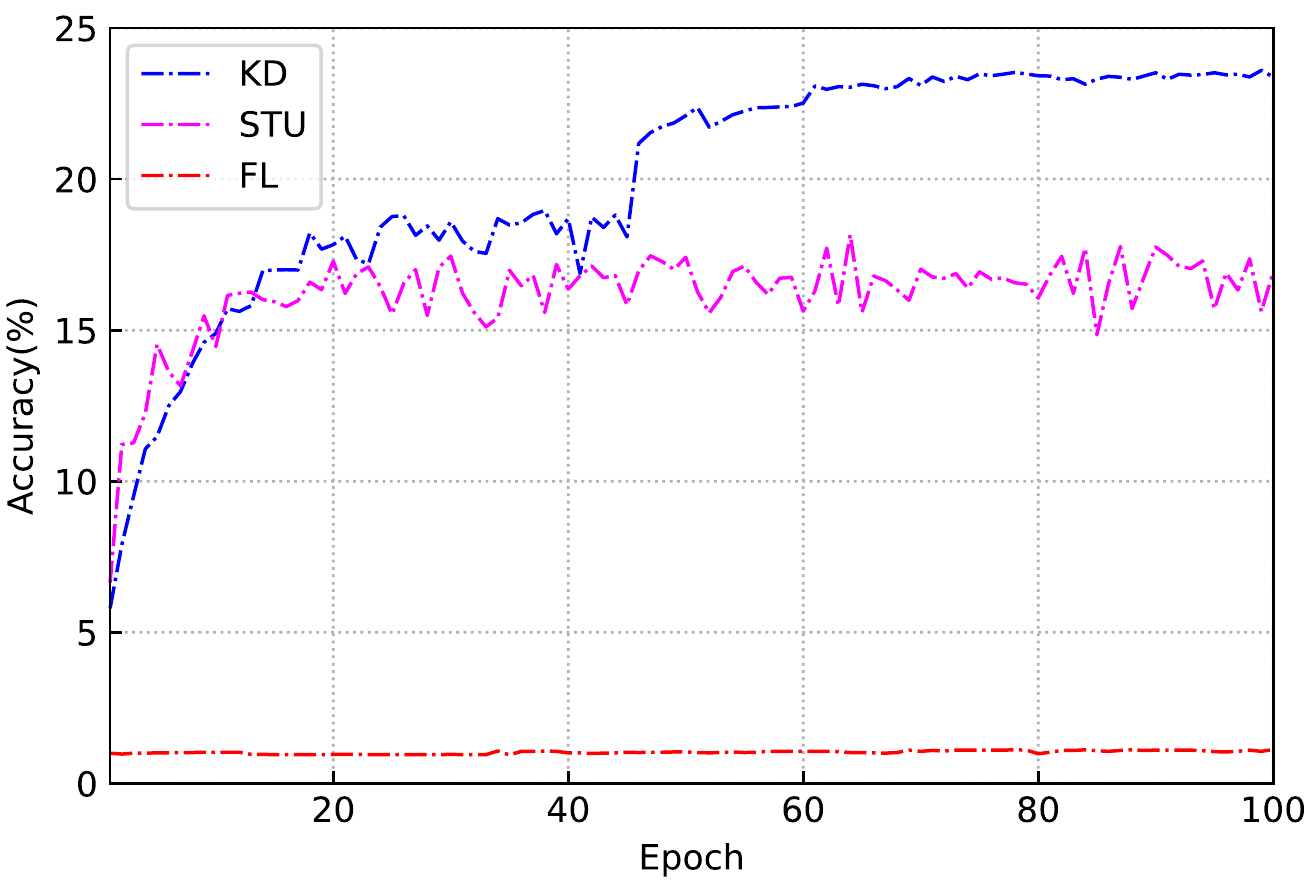}

  \caption {Full test set accuracy with Non-IID training set.}
  \label{fig4}

\end{figure}

\begin{figure}[h]
  \centering
  \includegraphics[width=0.9\columnwidth]{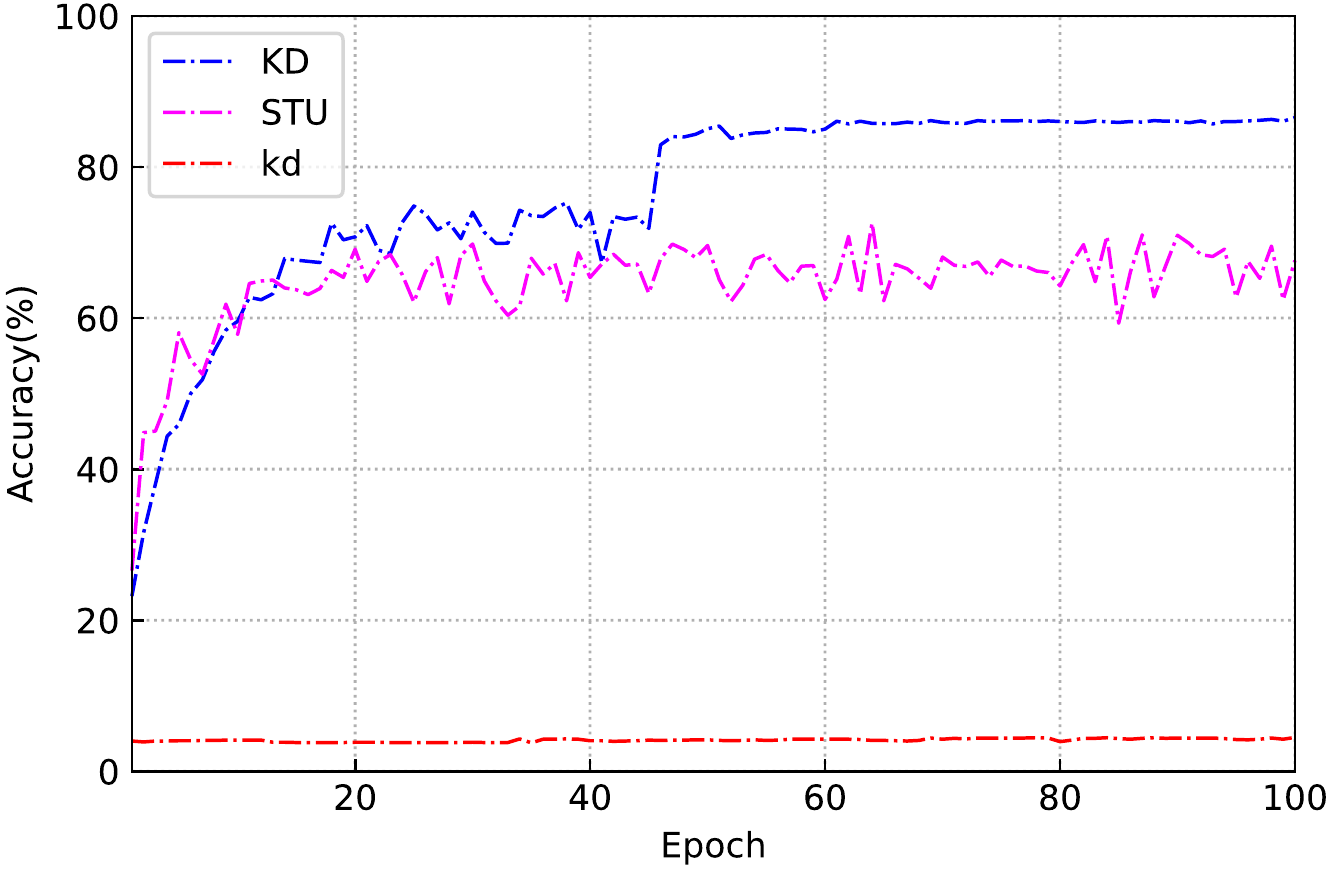}

  \caption {Private test set accuracy with Non-IID training set.}
  \label{fig5}

\end{figure}

\begin{figure}[h]
  \centering
  \includegraphics[width=0.9\columnwidth]{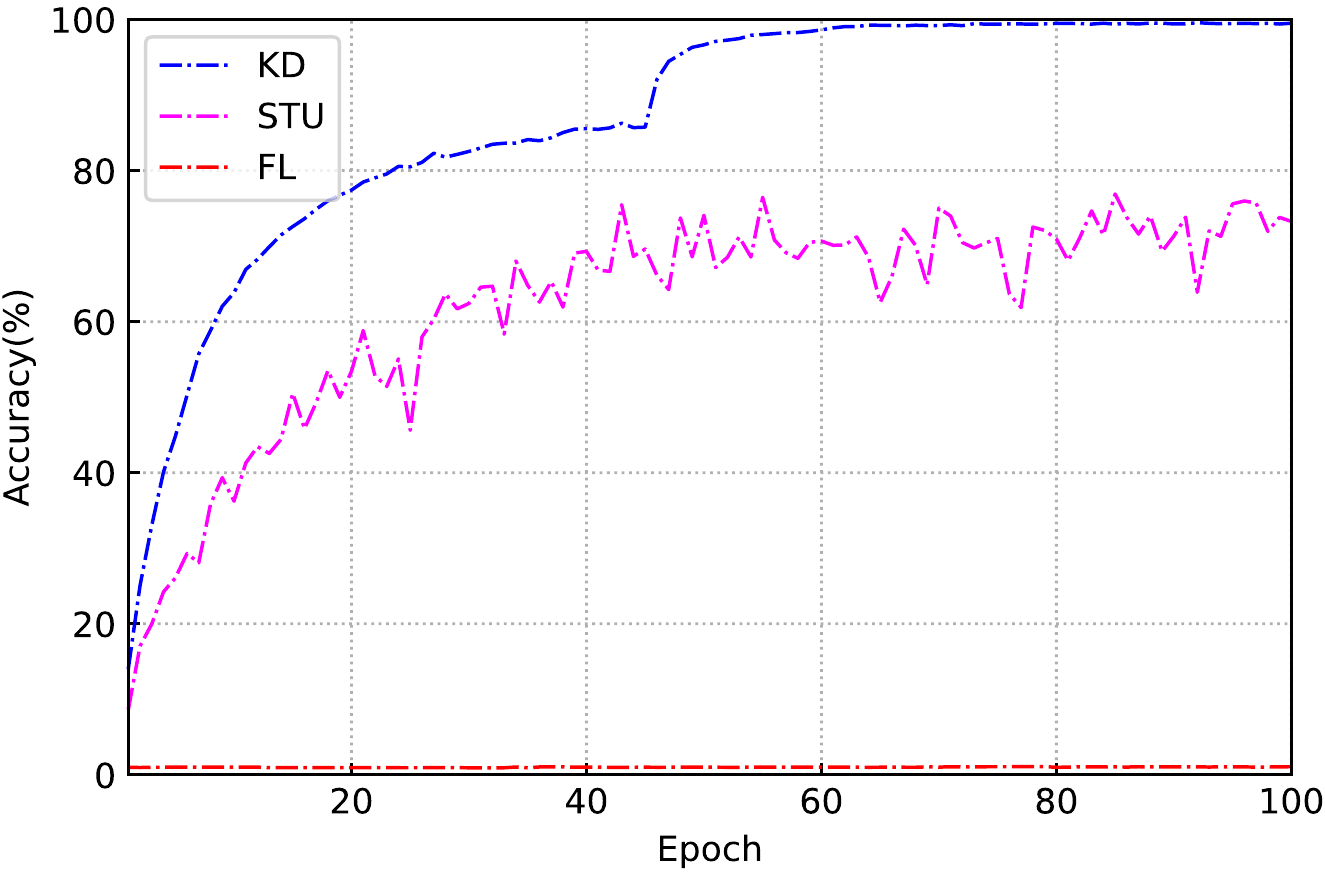}

  \caption {Train accuracy with Non-IID training set.}
  \label{fig6}

\end{figure}

Figure \ref{fig4} - Figure \ref{fig6} show the training results on the Non-IID datasets. To illustrate the detrimental effect of heterogeneous datasets on accuracy, we examine the accuracy of the model on the original test data and the heterogeneous test data, respectively. The original dataset contains all 100 kinds of test images, and the heterogeneous dataset only extracts 25 kinds of images that match the training data. Since the traditional FL needs to aggregate the parameters of the models from different clients, the accuracy of the model on the heterogeneous test set is comparable to that of the original test set. Obviously, FL fails to converge at all on the heterogeneous dataset. Even after 100 epochs of training, the classification accuracy is still around $1\%$, which is basically the same as that of random classification. This result is consistent with the findings of Sattler \emph{et al}.\cite{8889996}. Although the FL method is set up for a federation learning scenario, it only considers the IID data distribution among clients. In this case, the expectation of the empirical risk function across clients is usually considered to be an unbiased estimate of the expectation of the total data. The above results demonstrate that good performance can be achieved using the FL method. However, the unbiasedness in the above assumptions no longer holds in the Non-IID case. As a result, the weights of the model are updated in different directions for different clients during the training process. Since traditional FL uses weight averaging to aggregate the parameters of each model, the Non-IID data distribution makes it no longer convergent \cite{zhao2018federated}.

\begin{figure}[h]
  \centering
  \includegraphics[width=0.9\columnwidth]{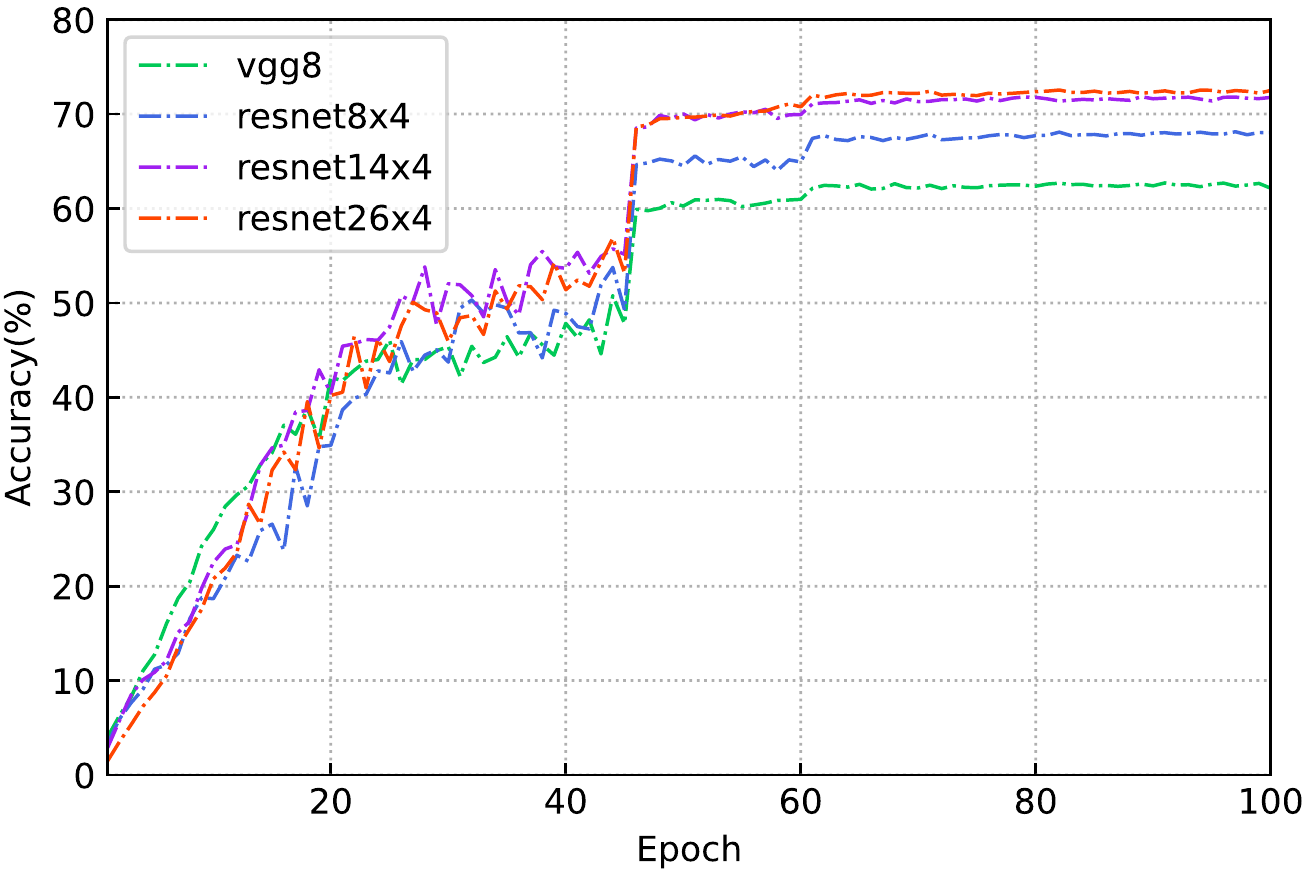}

  \caption {Test accuracy of different models training with KD on IID training set.}
  \label{fig7}

\end{figure}

\begin{figure}[h]
  \centering
  \includegraphics[width=0.9\columnwidth]{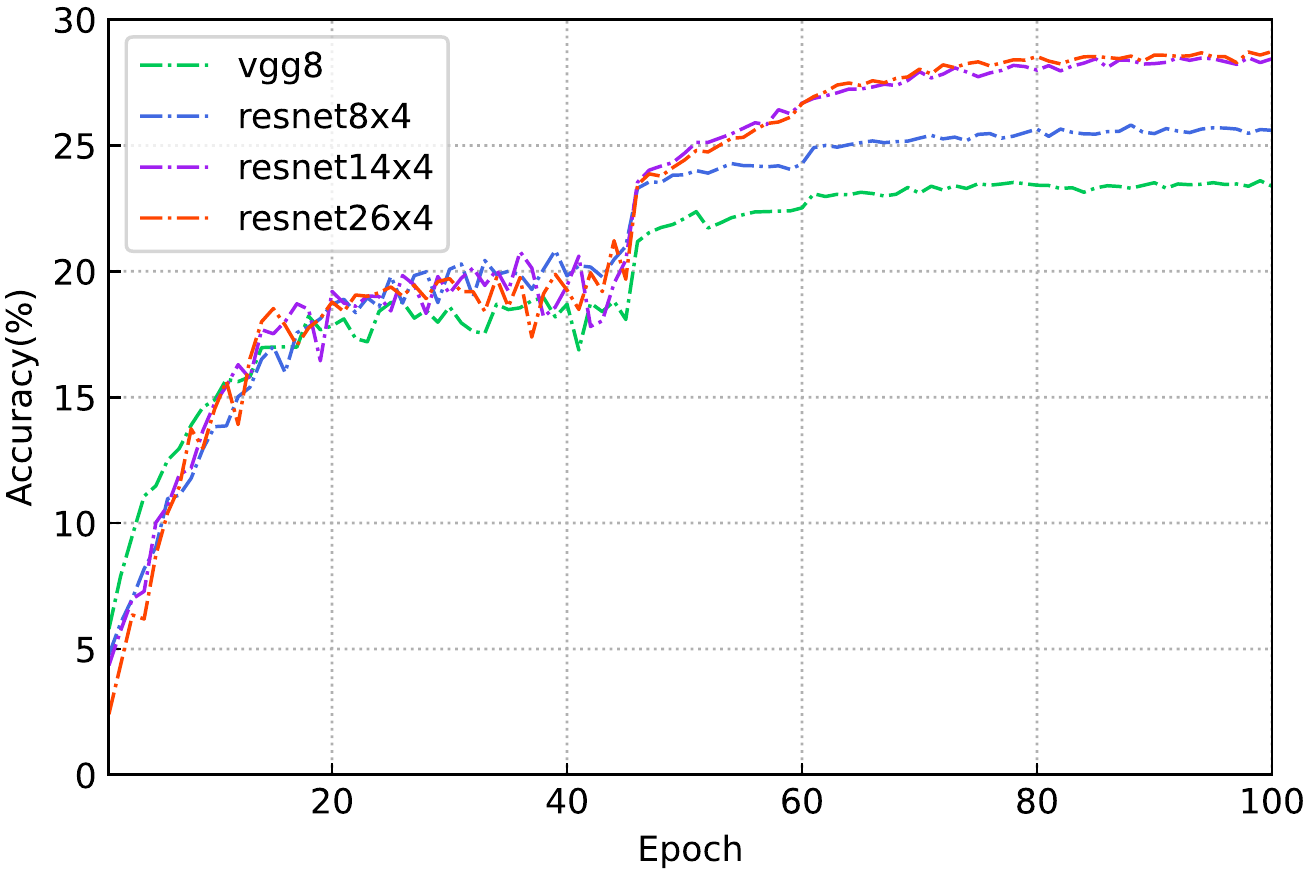}

  \caption {Test accuracy of different models training with KD on Non-IID training set.}
  \label{fig8}

\end{figure}




\begin{table*}[h]
  \centering
  \caption{Test Accuracy ($\%$) On The Full Dataset.}
    \begin{tabular}{c|c|c|c|c|c|c|c|c|c|c|c|c}
    \toprule[0.75pt]\toprule[0.75pt]
  {Method} & \multicolumn{4}{c|}{KD} & \multicolumn{4}{c|}{FL} & \multicolumn{4}{c}{Student Model} \\
    \hline
    \multirowcell{2}{Training set\\Model}&
   \multicolumn{2}{c|}{IID} & \multicolumn{2}{c|}{Non-IID}  & \multicolumn{2}{c|}{IID} & \multicolumn{2}{c|}{Non-IID} & \multicolumn{2}{c|}{IID} & \multicolumn{2}{c}{Non-IID}  \\

    \cline{1-13}
    &\multicolumn{1}{c|}{Top-1}&Top-5&\multicolumn{1}{c|}{Top-1}&Top-5&\multicolumn{1}{c|}{Top-1}&Top-5&\multicolumn{1}{c|}{Top-1}&Top-5&\multicolumn{1}{c|}{Top-1}&Top-5&\multicolumn{1}{c|}{Top-1}&Top-5\\

    \hline
            VGG-8&62.71&86.61&23.60&63.51&59.90&84.49&1.12&5.01&43.23&73.83&18.16&23.62\\ \hline
            ResNet-8x4&68.13&90.85&25.81&67.88&62.13&87.00&1.47&5.40&45.26&76.53&18.50&24.00\\ \hline
            ResNet-14x4&71.84&92.40&28.48&73.43&65.64&88.50&1.43&5.11&49.97&79.29&19.35&24.08\\ \hline
            ResNet-26x4&72.55&92.59&28.73&73.51&66.10&89.22&1.23&5.04&49.32&78.66&19.53&24.20\\ \hline
          \bottomrule[1.1pt]
    \end{tabular}
  \label{tab:1}
\end{table*}

\begin{table*}[h]
  \centering
  \caption{Test Accuracy ($\%$) On The Private Dataset.}
    \begin{tabular}{c|c|c|c|c|c|c}
    \toprule[0.75pt]\toprule[0.75pt]
  \multirowcell{2}{Method\\Model}&
  \multicolumn{2}{c|}{KD} & \multicolumn{2}{c|}{FL}  & \multicolumn{2}{c}{Student Model}   \\
    \cline{1-7}
    &\multicolumn{1}{c|}{Top-1}&Top-5&\multicolumn{1}{c|}{Top-1}&Top-5&\multicolumn{1}{c|}{Top-1}&Top-5\\
    \hline
            VGG-8&86.53&96.75&4.47&20.04&72.59&94.47\\ \hline
            ResNet-8x4&89.46&98.11&5.88&21.60&73.95&95.99\\ \hline
            ResNet-14x4&90.98&98.11&5.72&20.44&77.36&96.31\\ \hline
            ResNet-26x4&90.14&97.75&4.92&20.16&78.08&96.79\\
          \bottomrule[0.75pt]\bottomrule[0.75pt]
    \end{tabular}
  \label{tab:2}
\end{table*}

While the proposed KD-driven method shows good adaptability on the Non-IID dataset. Unlike traditional knowledge distillation schemes, simKD aims at making the student model learn the feature extraction capabilities of the teacher model. In the Non-IID case, the student model has only 25 training labels, meaning that the labels of the remaining 75 images are always zero for the classifier. This leads the model to reduce the classification probability of these 75 images as much as possible during training in order to obtain higher training accuracy. Therefore, it leads to a significant decrease in the accuracy of the model on the complete test data. This problem is largely alleviated by making the student model reuse the classifier of the teacher model and perform deep feature alignment. This is because in image recognition tasks, the operations performed to extract features from different images are often similar, while the classifier must learn based on a specific image class. The accuracy of the training set illustrates that the KD-driven method can converge quickly and has good stability. It can be seen that the accuracy on the heterogeneous test set is significantly higher than that on the original dataset, which is due to the fact that dividing the dataset keeps the distribution of test and training data consistent, while the class and total number of images are greatly reduced. Therefore, the training difficulty of the model decreases. In this case, KD is able to achieve a $14\%$ performance improvement compared to STU. The accuracy of both KD and STU decreases significantly on the original test set. Clearly, the lack of 75 classes of training images severely affects the classification ability of the model, and the huge difference between the results in Figure \ref{fig4} and Figure \ref{fig5} indicates that the model has difficulty identifying the missing classes of images in the training set. In this case, learning the knowledge of the teacher model can effectively compensate for the adverse effects of missing training data, and a $5.4\%$ performance improvement is obtained compared to STU.

KD not only improves the performance of client models, but also adapts well to different sizes of student models.
Figure \ref{fig7} and Figure \ref{fig8} show the tested accuracy of different client-side models trained with KD. It can be observed that each model can converge quickly and stabilize to a high accuracy. The final accuracy achieved by different models varies, which is determined by the representational power of the models themselves.
In addition, the models converge faster when trained on IID data than Non-IID data. This is because the number of images is the same for both data, and the difference in the distribution between the heterogeneous training sets and test sets increases the difficulty of training. 

Table \ref{tab:1} and Table \ref{tab:2} show the optimal accuracy of all methods to train different client models. It is clear that KD achieves the optimal performance in all cases. In contrast, FL is able to achieve good performance on IID data, but cannot handle Non-IID data. While STU does not suffer from the same catastrophic accuracy degradation as FL, its performance is also unsatisfactory on Non-IID data.

\subsection{Performance of joint optimization of Q-learning and convex optimization}

\begin{table*}[h]
    \centering
    \caption{Performance of Different Algorithms on Non-IID Dataset.}

	\begin{tabular}{c|c|c|c|c}
		\toprule[0.75pt]\toprule[0.75pt]
Metric        & Proposed Method &Q-learning Only  & FL-Min  &FL-Max       \\ \hline
Objective Value    & -1.03 &-0.99  & -0.95 & 0.0075 \\ \hline
Average Delay      & 14.33 s &15.66 s & 14.02 s & 27.01 s \\ \hline
Acc-Own            & 90.19 \% & 89.72 \%& 1.12 \% &1.23 \% \\ \hline
Acc-Avg            & 27.10 \% & 26.52 \%& 1.12 \% &1.23 \% \\ \hline
Probability of VGG-8 & 1.5 \% & 6.5 \% & 100 \% & 0 \%       \\ \hline
Probability of ResNet-8x4 & 50.5 \% & 62.0 \% & 0 \% & 0 \%       \\ \hline
Probability of ResNet-14x4 & 48.0 \% & 28.0 \% & 0 \% & 0 \%       \\ \hline
Probability of ResNet-26x4 & 0.0 \%& 3.5 \% & 0.0 \% & 100 \%       \\ 
		\bottomrule[0.75pt]\bottomrule[0.75pt]
	\end{tabular}
	\label{tab:3}
\end{table*}

\begin{table*}[!ht]
    \centering
    \caption{Performance of Different Algorithms on IID Dataset.}

	\begin{tabular}{c|c|c|c|c}
		\toprule[0.75pt]\toprule[0.75pt]
Metric             & Proposed Method &Q-learning Only         & FL-Min  &FL-Max       \\ \hline
Objective Value    & -1.27   & -1.22 &-1.11 & 0.0074 \\ \hline
Average Delay      & 13.91 s &18.09 s & 13.07 s & 20.55 s \\ \hline
Acc-Own            & 71.17 \% &70.47 \% & 59.9 \% &65.64 \% \\ \hline
Acc-Avg            & 71.17 \% &70.47 \% & 59.9 \% &65.64 \% \\ \hline
Probability of  VGG-8 & 0.5 \% &0.0 \% & 100 \% & 0 \%       \\ \hline
Probability of  ResNet-8x4 & 17.0 &39 \% & 0 \% & 0 \%       \\ \hline
Probability of  ResNet-14x4 & 80.0&49.5 \% & 0 \% & 0 \%       \\ \hline
Probability of  ResNet-26x4 & 2.5 \% &11.5 \% & 0.0 \% & 100 \%       \\ 
		\bottomrule[0.75pt]\bottomrule[0.75pt]
	\end{tabular}
	
	\label{tab:4}
\end{table*}

To evaluate the performance of the proposed joint optimization of Q-learning and convex optimization, we compare it with the other three methods. The Q-learning only uses the Q-learning to select student models and allocate resources for all users, by comparing with this method the effectiveness of jointly using convex optimization can be proved. The FL-Min and FL-Max simply choose the same minimum/maximum student models for all users and then use the Q-learning to determine the offloading method and convex optimization to allocate resources, by comparing these two methods the importance of selecting a different model for different users can be shown. In Table \ref{tab:3} and Table \ref{tab:4}, the average delay is the average time taken to train an epoch by $K$ users in the system, and the Acc-Own is the accuracy of the $K$ student models on its own private dataset, and the Acc-Avg is the average accuracy of the $K$ student model on all users' dataset. We also counted the probability of different student models being selected when using different algorithms, as is shown in the last four lines in Table \ref{tab:3} and Table \ref{tab:4}. As is shown in Table \ref{tab:3} and Table \ref{tab:4} the proposed joint optimization methods always achieve the best performance in objective value, because Q-learning allows users to choose the best student model and offload methods, and convex optimization can find the best resource allocation solution on the basis of Q-learning. Moreover, according to \cite{watkins1992q}, in problems with finite discrete action spaces, Q-learning always converges to the Bellman optimal solution after adequate training epochs. Since both $m$ and $x$ are discrete values, the Q-learning can achieve the Bellman optimal solution, theoretically. However, if Q-learning is used to jointly select student models and allocate resources, the action space of the problem will be infinite, since $f$ and $b$ are continuous values. Therefore, it is challenging for Q-learning to achieve high performance, and the Q-learning only method always has lower performance than proposed joint optimization methods. Remarkably, although the Q-learning only method is worse than joint optimization, due to the resources management, it is always better than FL-Max and FL-Min with the best resources allocation scheme. This phenomenon reveals the importance of choosing different student models for devices with heterogeneous computing resources. The FL-Min always has a minimum average delay in training one epoch due to the low transmission delay and training delay caused by the minimum model size.

According to Table \ref{tab:1} and Table \ref{tab:2}, although the larger student model has a higher average performance in all users' datasets, the performance of the largest model ResNet-26x4 on the user's private data is not better than ResNet-14x4. This is because large models have a greater ability to extract general features of the dataset, as a result, its performance on a specific dataset may not be as good as that of a small model due to the effect of average features. Due to the heterogeneous objectives of the users, the student model needs to achieve good performance on the user's own private dataset, thus the Q-learning-based model selection algorithm rarely selects the largest ResNet-26x4 as the student model since the its accuracy on user's private dataset is poor.

\section{Conclusion} \label{sec-5}
In this paper, we have studied the problem of heterogeneity of data, computing resources, and objectives for users in FL systems. We have proposed a DT-assisted KD framework so that different users can select different NN models in FL systems. Then, we have jointly used Q-learning and convex optimization to select NN models and allocate resources for each user. Simulation results have validated the effectiveness and efficiency of the proposed framework and joint optimization method. Our work can offer valuable insights into the importance yet the under-explored field of heterogeneity in FL systems, and give a promising application for DT technology. In future work, we will focus on the distributed optimization method to solve the problem of model selection and resource management.

\section*{Acknowledge}
This work was supported by the National Key Research and Development Program of China (2020YFB1807700), the National Natural Science Foundation of China (NSFC) under Grant No. 62071356, the Chongqing Key Laboratory of Mobile Communications Technology under Grant cqupt-mct-202202.

\bibliographystyle{gbt7714-numerical}
\bibliography{ref}
\biographies

\begin{CCJNLbiography}{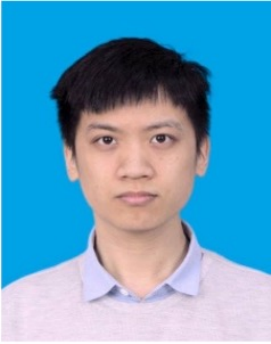}{Xiucheng Wang}
received a B.S. degree in telecommunication engineering from Xidian University in 2021, and is currently pursuing an M.S. degree from Xidian University. His research area of
interest is digital twin and graph neural networks of the wireless network.
\end{CCJNLbiography}

\begin{CCJNLbiography}{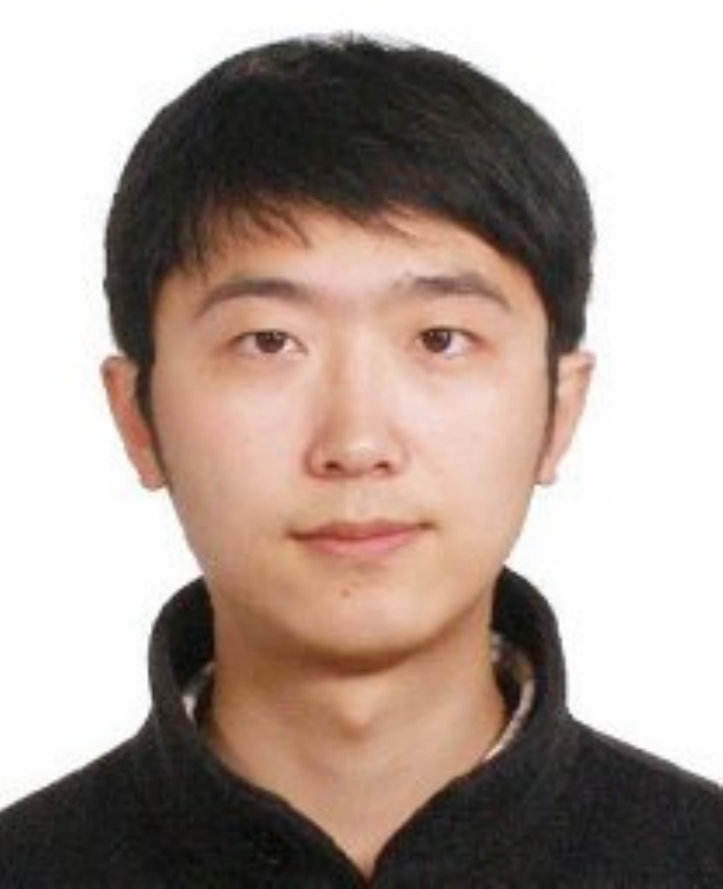}{Nan Cheng}
Nan Cheng received the Ph.D. degree from the Department of Electrical and Computer Engineering, University of Waterloo, Waterloo, ON, Canada, in 2016, and the B.E. and M.S. degrees from the Department of Electronics and Information Engineering, Tongji University, Shanghai, China, in 2009 and 2012, respectively. \\
He worked as a Postdoctoral Fellow with the Department of Electrical and Computer Engineering, University of Toronto, Toronto, ON, Canada, from 2017 to 2019. He is currently a Professor with the State Key Laboratory of Integrated Services Networks and the School of Telecommunications Engineering, Xidian University, Xi’an, Shaanxi, China. His current research focuses on B5G/6G, space–air–ground-integrated network, big data in vehicular networks, and self-driving system. His research interests also include performance analysis, MAC, opportunistic communication, and application of AI for vehicular networks.

\end{CCJNLbiography}
\begin{CCJNLbiography}{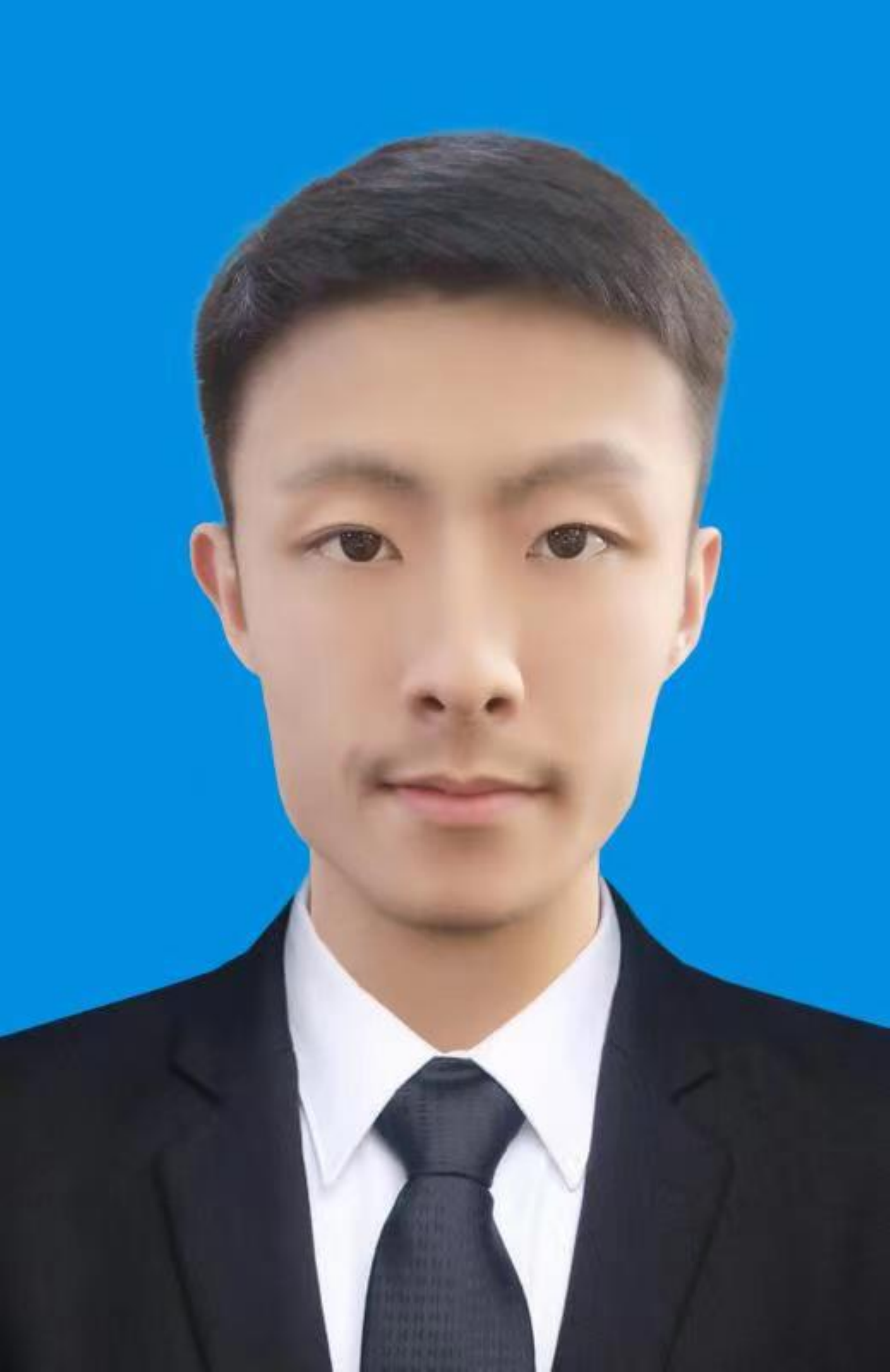}{Longfei Ma}
received a B.S. degree in telecommunication engineering from Xidian University in 2022, and is currently pursuing an M.S. degree from Xidian University. His research area of
interest is digital twin and dynamic neural network of the wireless network.
\end{CCJNLbiography}

\begin{CCJNLbiography}{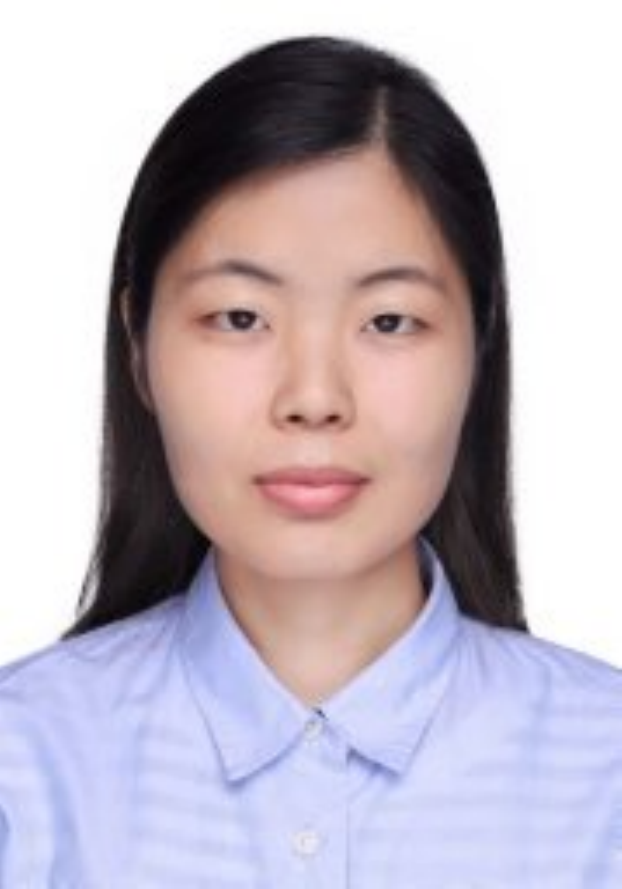}{Ruijin Sun}
received the Ph.D. degree from the Beijing University of Posts and Telecommunications, China, in 2019. She is currently a Lecturer at the School of Telecommunications Engineering, and with State Key Lab of ISN, Xidian University, Shanxi, China. She worked as a joint Postdoctoral Fellow with Peng Cheng Laboratory and Tsinghua University from 2019 to 2021. She was a Visiting Student at University of Waterloo, Canada (September 2017-September 2018). Her research interests are in the area of knowledge-driven wireless resource allocation and MIMO signal processing.
\end{CCJNLbiography}

\begin{CCJNLbiography}{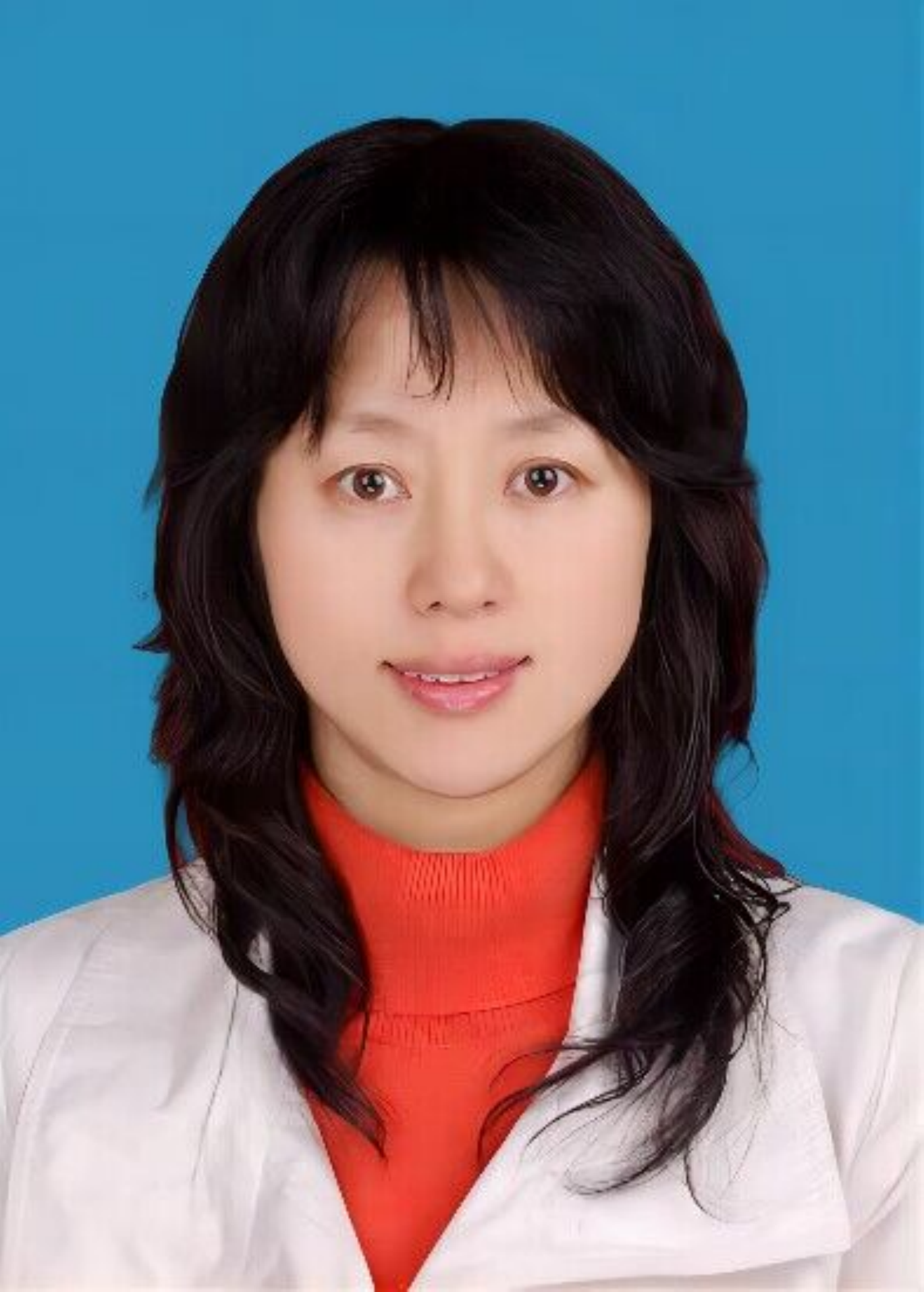}{Rong Chai}
received the B.E. and M.S. degrees from the University of Electronic Science and Technology of China, Chengdu, China, in 1995 and 1998, respectively, and the Ph.D. degree in electrical engineering from McMaster University, Hamilton, ON, Canada, in 2008. In 2008, she joined the School of Communication and Information Engineering, Chongqing University of Posts and Technology, Chongqing, China, where she is currently a Professor. Her research interest is in wireless communication and network theory.
\end{CCJNLbiography}

\begin{CCJNLbiography}{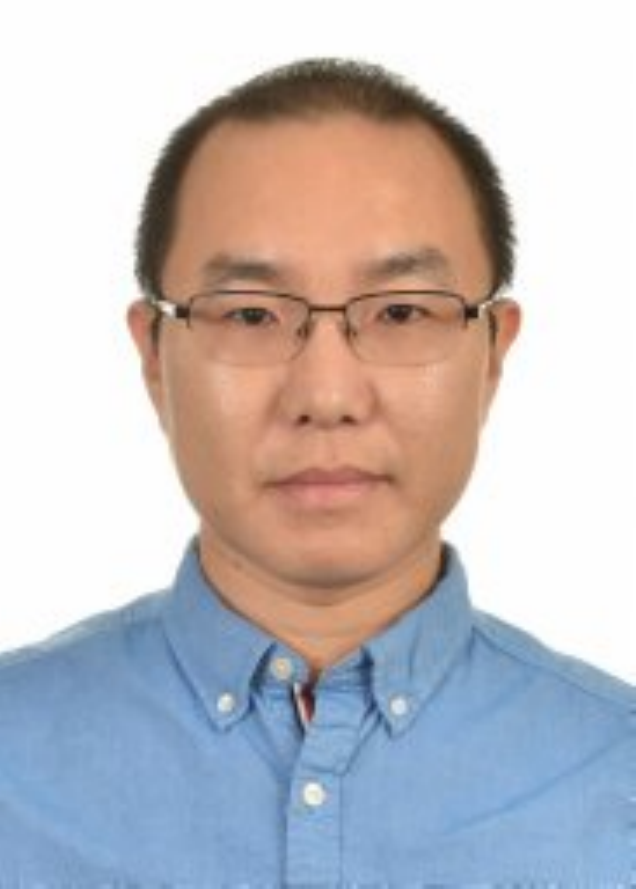}{Ning Lu}
received the B.Eng. and M.Eng. degrees in electrical engineering from Tongji University, Shanghai, China, in 2007 and 2010, respectively, and the Ph.D. degree in electrical engineering from the University of Waterloo, Waterloo, ON, Canada, in 2015. He is currently an Assistant Professor with the Department of Electrical and Computer Engineering, Queen’s University, Kingston, ON, Canada. Prior to joining Queen’s University, he was an Assistant Professor with the Department of Computing Science, Thompson Rivers University, Kamloops, BC, Canada. From 2015 to 2016, he was a Postdoctoral Fellow with the Coordinated Science Laboratory, University of Illinois at UrbanaChampaign, Champaign, IL, USA. He also spent the summer of 2009 as an Intern with the National Institute of Informatics, Tokyo, Japan. He has authored or coauthored more than 40 papers in top IEEE journals and conferences, including the IEEE/ACM TRANSACTIONS ON NETWORKING, IEEE JOURNAL ON SELECTED AREAS IN COMMUNICATIONS, ACM MobiHoc, and IEEE INFOCOM. His current research interests include real-time scheduling, distributed algorithms, and reinforcement learning for wireless communication networks.
\end{CCJNLbiography}

\end{document}